\pdfoutput=1

\documentclass{article}
\usepackage{ijcai21}

\usepackage{times}

\usepackage{soul}
\usepackage{url}
\usepackage[utf8]{inputenc}
\usepackage[small]{caption}
\usepackage{graphicx}
\usepackage{amsmath}
\usepackage{booktabs}
\usepackage{float}
\usepackage{bm}
\usepackage{amsfonts}
\usepackage[numbers]{natbib}
\usepackage[dvipsnames,table,xcdraw]{xcolor}
\usepackage{subcaption}
\usepackage{graphicx}
\usepackage{wrapfig}

\usepackage{multirow}
\usepackage{graphicx}
\usepackage{placeins}
\usepackage{booktabs,makecell,tabularx}

\usepackage[hang,flushmargin]{footmisc}

\title{Enhancing Topic Extraction in Recommender Systems with Entropy Regularization}

\author{
Xuefei Jiang \and
Dairui Liu\and
Ruihai Dong
\affiliations
School of Computer Science, University College Dublin, Dublin, Ireland
\emails
\{xuefei.jiang, dairui.liu\}@ucdconnect.ie, ruihai.dong@ucd.ie
}

\begin{document}
\maketitle

\begin{abstract}
In recent years, many recommender systems have utilized textual data for topic extraction to enhance interpretability. However, our findings reveal a noticeable deficiency in the coherence of keywords within topics, resulting in low explainability of the model. This paper introduces a novel approach called entropy regularization to address the issue, leading to more interpretable topics extracted from recommender systems, while ensuring that the performance of the primary task stays competitively strong. The effectiveness of the strategy is validated through experiments on a variation of the probabilistic matrix factorization model that utilizes textual data to extract item embeddings. The experiment results show a significant improvement in topic coherence, which is quantified by cosine similarity on word embeddings. The source code is made available here.\footnote{\url{https://github.com/ScXfjiang/conv_pmf}}
\end{abstract}

\section{Introduction}
Recommender systems have gained significant interest across numerous fields including e-commerce, film streaming, and social networking, due to their ability to predict user preferences \cite{bobadilla2013recommender}. As a common way to build recommender systems, collaborative filtering (CF) generates recommendations or predicts unknown preferences for users by leveraging the preferences of other users who display similar tastes or behaviors \cite{su2009survey}. Among the plethora of collaborative filtering (CF) techniques, the latent factor model stands out by encoding users and items into vectors within an $n$-dimensional space, which are then employed to predict preferences or generate recommendations. Whereas traditional CF algorithms largely focus on user-item interactions like numerical ratings or click behavior, a growing number of models developed recently manage to harness textual data, for example, customer reviews \cite{kim2016convolutional,lu2018coevolutionary}, to mitigate such as data sparsity issues, which not only improve recommendation performances, but also provide opportunities to mine the meanings of latent features of CF models. 

Linking latent features with their semantic meanings can improve the interpretability of recommender systems \cite{srifi2020recommender}. Typically, topic modeling techniques are utilized to extract topics from textual data and align extracted topics to latent features. For example, \cite{mcauley2013hidden} introduces an approach that combines the traditional latent factor model with Latent Dirichlet Allocation (LDA) \cite{blei2003latent} to uncover topics correlated with the latent factors of both products and users. On the other hand, \cite{lu2018convolutional} adopts a convolutional neural network (CNN) to encode textual reviews into item embeddings. The convolutional kernels are then utilized to extract topics that correspond to the latent factors of items. However, for topic-based latent factor recommender systems, one critical issue that has emerged is the low coherence of keywords within extracted topics, which affects the interpretability of these topics and, consequently, the overall explainability of the recommender systems. The issue originates from the tendency of a single latent factor to assign equivalent weights to diverse semantics. Ideally, to enhance interpretability, each latent factor should encapsulate a single, unambiguous semantic theme.

This paper presents a novel optimization method called entropy regularization to address this challenge. The core idea is to drive each latent factor to concentrate on fewer aspects of textual reviews by introducing a regularization term, thereby amplifying the coherence of keywords associated with topics. Through experiments on a variation of the probabilistic matrix factorization model that utilizes textual data, we demonstrate that the proposed approach significantly improves topic coherence without compromising the performance of the recommender system.

\section{Related Work}
Latent factor models, a dominant subset of collaborative filtering techniques, are widely used in recommender systems. These models operate by encoding both users and items into fixed-length embeddings, thereby capturing the essential characteristics of each entity in a compact form. Among these models, matrix factorization is recognized for its simplicity and flexibility and is broadly accepted as the standard model in this category \cite{koren2009matrix}. In matrix factorization models, the rating matrix can be reconstructed by multiplying user and item embeddings. For instance, probabilistic matrix factorization, a classic latent factor model introduced in \cite{mnih2007probabilistic}, learns user and item matrices by solving an optimization problem that minimizes the discrepancy between actual and predicted rating values.

As the field continues to evolve, there is a growing demand for increasingly sophisticated models that leverage not just observed ratings, but also implicit feedback like purchase history and textual data. Several autoencoders have been proposed to encode a variety of implicit feedback into user or item embeddings. Among these autoencoders, neural networks stand out due to their potent representational capacity and flexibility. For example, \cite{elkahky2015multi} proposed a deep neural network structure that employs observed ratings, web browsing history, and search queries to enhance model performance. \cite{van2013deep} uses a convolutional neural network to predict latent factors from audio signals, addressing the cold start problem in music recommendation.

Later, people are not only satisfied by the improvement of the recommendation performance but also want to find out what is going on behind latent factor models. Among these various strategies, topic modeling techniques provide a compelling solution. These methods associate latent factors with a specific set of topic keywords, thereby offering an intuitive approach to interpreting latent factors. For instance, some studies adopt a combined approach, harnessing both numerical rating data and textual review data to create more accurate and interpretable recommender systems \cite{bao2014topicmf,cheng2018aspect,ling2014ratings,mcauley2013hidden}. In addition, attention mechanisms \cite{bahdanau2014neural, vaswani2017attention} are also employed to elucidate the semantics of latent factors by selectively focusing on parts of the input data. For example, \cite{liu2022novel} introduces a bi-level attention architecture to extract topic keywords and associate them with latent factors, thereby enhancing the interpretability of news recommendation systems.

\section{Methodology}
This section describes the base model of our study, convolutional matrix factorization (ConvMF), which is a latent factor model that utilizes textual reviews to generate item embeddings. In addition, we introduce our optimization technique, entropy regularization, which is designed to improve topic extraction within the base model.

\subsection{Convolutional Matrix Factorization}

\begin{figure}[ht]
    \centering
    \includegraphics[scale=0.3]{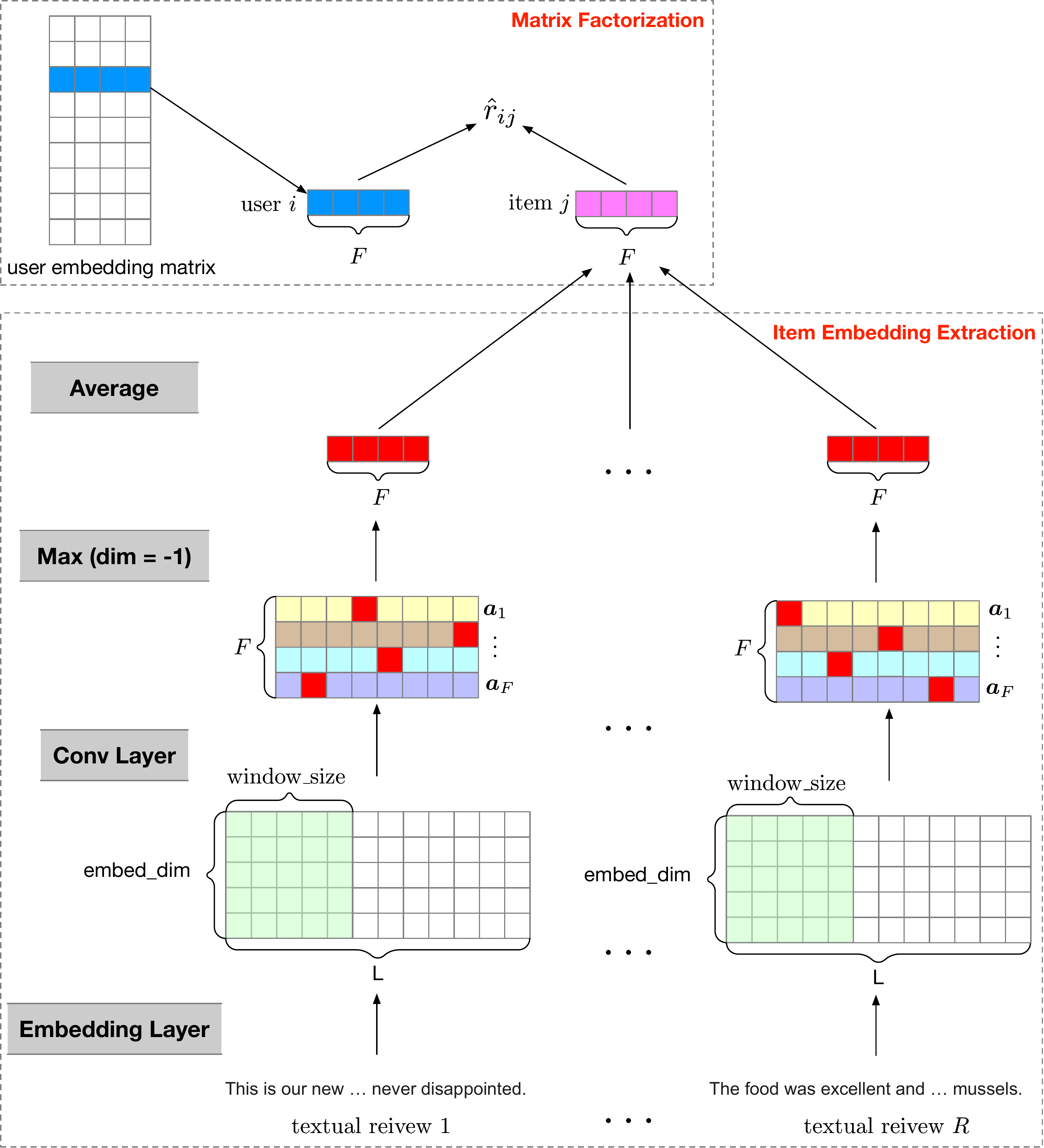}
    \caption{Structure of ConvMF. In this case, item $j$ is associated with $R$ textual reviews. The number of latent factors is denoted by $F$, which is also equivalent to the number of filters utilized in the convolutional layer.}
    \label{fig:conv_pmf}
\end{figure}

The base model used for the rating prediction task, as depicted in Figure \ref{fig:conv_pmf}, borrows the idea from convolutional matrix factorization \cite{lu2018convolutional}. In ConvMF, rather than creating and learning an item embedding matrix independently, the model employs a convolutional neural network to extract item embeddings from textual reviews. For each item embedding, we gather corresponding historical textual reviews and convert them into word embeddings. These word embeddings are then processed through the convolutional neural network, and the maximum activation values of each convolutional filter are identified. The final item embedding is obtained by averaging the embeddings extracted from all reviews associated with the item. However, the user embeddings follow the traditional probabilistic matrix factorization \cite{mnih2007probabilistic}; they are assembled into a user embedding matrix that is initialized and learned from scratch. Ultimately, the estimated user rating for an item is achieved by taking the dot product of the user embedding and item embedding.

\subsection{Topic Extraction}
Beyond simply predicting ratings, ConvMF also endeavors to identify the semantic meaning of each latent factor. Once training concludes, we can loop over all textual reviews within the training set and leverage the trained convolutional kernel to compute activations for all n-grams. When calculating the activation values for individual words, we assume that the activation value is uniformly distributed across each word in the n-gram. By determining the average activation values for each word, we can identify the top-$k$ words with the highest activation values for each latent factor, which serve to define the latent factor's interpretation. From a topic modeling perspective, these highlighted keywords can be understood as the underlying topic of the corresponding latent factor.

\begin{figure}[ht]
\centering
\includegraphics[scale=0.3]{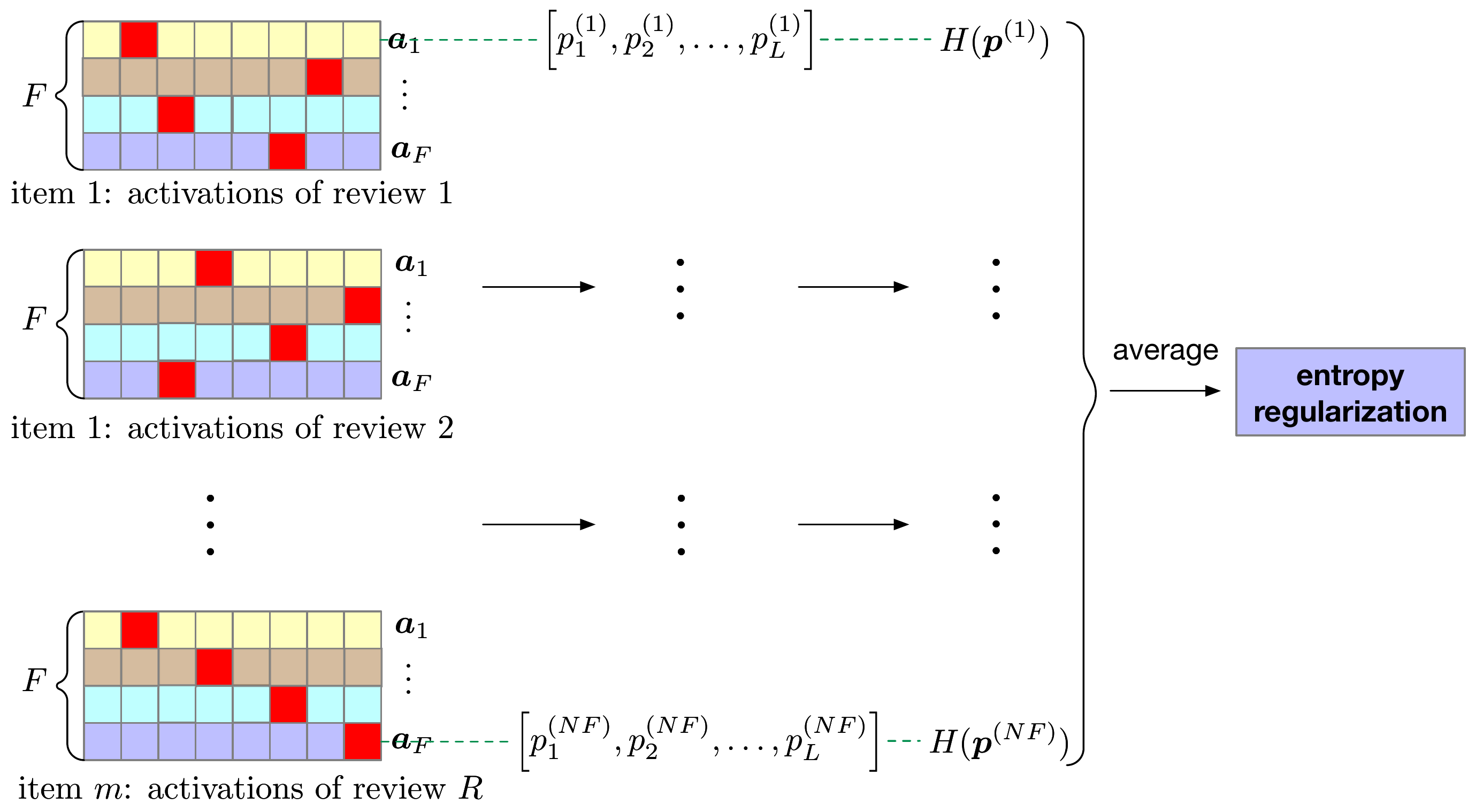}
\caption{Computation of the entropy regularization term across a batch consisting of $m$ items. In this context, $N$ is the total count of textual reviews within the batch, $F$ is the number of latent factors, and $L$ signifies the length of textual reviews. Total $N\times F$ entropy is computed and averaged to yield the final entropy regularization term.}
\label{fig:entropy_regularizatioin}
\end{figure}

\subsection{Entropy Regularization}
Ideally, we would like each latent factor to possess a single, unambiguous semantic meaning. For instance, in a food dataset, the first latent factor might correspond to keywords related to \textit{taste}, the second to keywords related to \textit{price}, the third to keywords related to \textit{location}, and so forth. However, we often observe that the activation values of keywords within a single latent factor are remarkably similar. This pattern suggests that the latent factor assigns comparable weight to different semantic meanings, which subsequently leads to reduced keyword coherence within each latent factor and an ambiguous interpretation of the latent factor. For instance, a latent factor might concurrently assign similar and low weights to varied semantics such as \textit{taste}, \textit{price}, and \textit{location}.

To address this issue, we employ entropy regularization to force each latent factor to concentrate on fewer aspects of textual reviews. As depicted in Figure \ref{fig:entropy_regularizatioin}, each convolutional filter processes textual reviews and generates corresponding activations. For a review activation $ \bm{a}\in\mathbb{R}^L $, where $ L $ is the length of the review, we employ a \textit{softmax} function to transform it into a probability distribution

\begin{equation}
    \bm{p} = softmax\left(\bm{a}\right)
\end{equation}
where
\begin{equation}
    p_i = \frac{e^{a_i}}{\sum_{j} e^{a_j}}\text{,}
\end{equation}

\noindent
and then calculate its information entropy
\begin{equation}
    H(\bm{p}) = -\sum_{i} p_i \log_2 p_i\text{.}
\end{equation}

The quantity $H(\bm{p})$ provides insight into how the model distributes weights across the corresponding textual review. Specifically, $H(\bm{p})$ is large when the weights are evenly distributed among words, while it is small when the weights are concentrated on a few words and assigned low values on most other words. Within a given batch, the entropy of all textual reviews is calculated, averaged, and subsequently introduced to the loss function as a regularization term, 

\begin{equation}
    Loss = RMSE + \lambda\cdot\frac{1}{NF}\sum_{i=1}^{NF} H(\bm{p}^{(i)})\text{,}
\end{equation}

\noindent
where $N$ is the total number of reviews within a batch, $F$ is the number of latent factors, and $\lambda$ is the coefficient of the entropy regularization term.

\begin{figure}[ht]
\begin{subfigure}{\linewidth}
    \centering
    \includegraphics[width=0.8\linewidth]{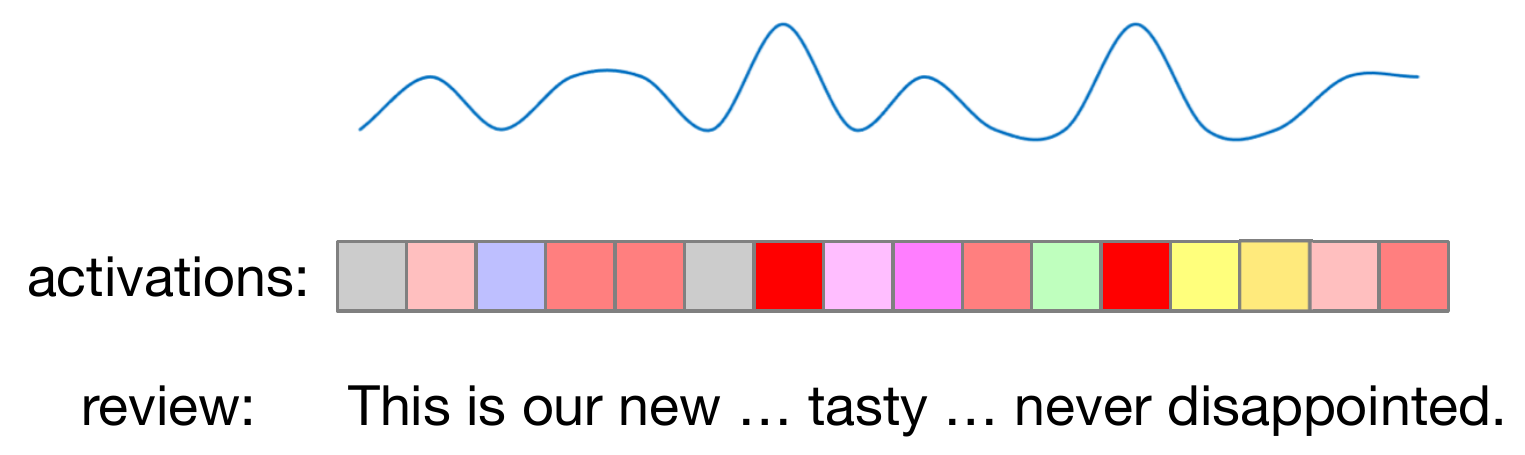}\hfill
    \subcaption{Activation values with high information entropy.}
\end{subfigure}\par\medskip
\begin{subfigure}{\linewidth}
    \centering
    \includegraphics[width=0.8\linewidth]{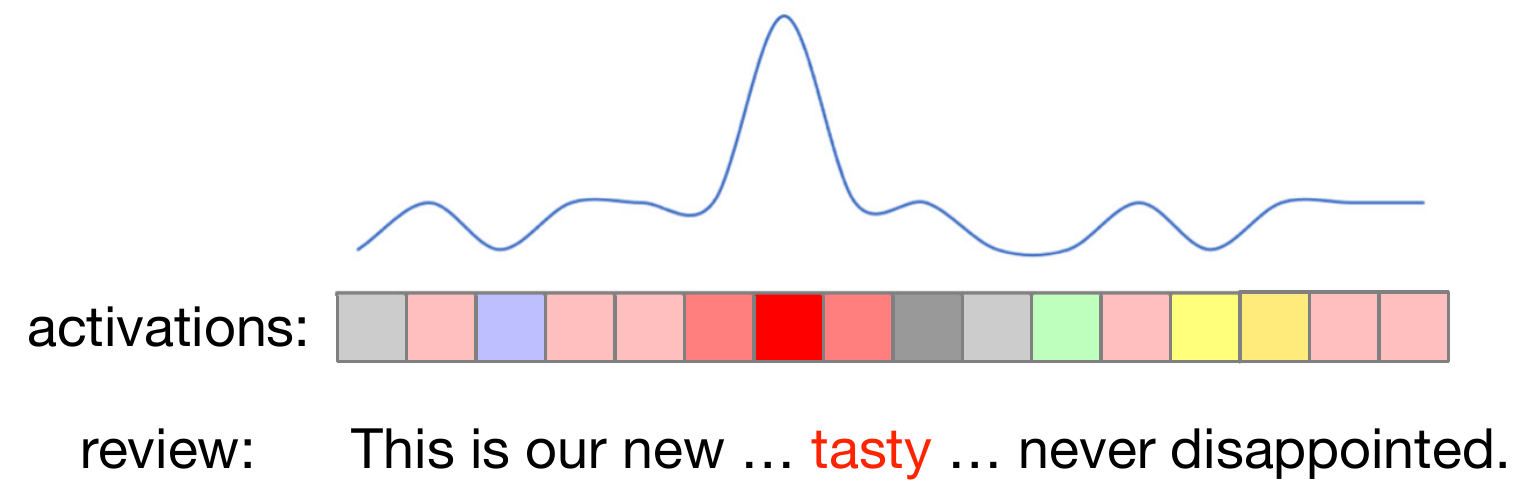}\hfill
    \subcaption{Activation values with low information entropy.}
\end{subfigure}\par\medskip
\caption{Depicting the impact of entropy regularization on activation value distribution. Subfigure (a) shows the distribution of activation values within a textual review prior to the application of entropy regularization, while subfigure (b) illustrates the change in the same distribution after entropy regularization has been introduced.}
\label{fig:function_of_entropy}
\end{figure}

This entropy regularization term affects the distribution of activation values of textual reviews and serves to penalize reviews that exhibit uniformly distributed activation values. Figure \ref{fig:function_of_entropy} showcases the effect of entropy regularization on the distribution; subfigures (a) and (b) illustrate the activation values for the same textual review but with high and low information entropy, respectively. By comparing the two subfigures, we can see that the model in (a) gives similar attention to each word in the review, whereas the model in (b) focuses on the word \textit{tasty}. Thus, by promoting activation values with lower information entropy, entropy regularization compels the filter to concentrate on fewer but more semantically significant aspects of the review.

\section{Experiments}
This section presents an evaluation of ConvMF enhanced by entropy regularization, assessing its performance in terms of both topic coherence and rating prediction accuracy.

\paragraph{Datasets} The evaluation of our model is performed on the Amazon Grocery and Gourmet Food dataset\footnote{\url{https://nijianmo.github.io/amazon/index.html}}. This dataset is a part of the Amazon review datasets which includes both textual reviews and numerical ratings. The Amazon Grocery and Gourmet Food dataset is selected due to its relatively appropriate number of consumers and items. In addition, this dataset also contains sufficient textual reviews for our model to accurately extract item embeddings from these textual reviews. Table \ref{tab:statistics_amazon_food} provides key statistics of the dataset.
%We employ the $5$-core version of the dataset to ensure that each user or item has at least $5$ textual reviews and numerical ratings.

\begin{table}[H]
\centering
    \begin{tabular}{ll}
    \hline
    statistic & value \\
    \hline
    $\#$ users                         & $14,681$     \\
    $\#$ items                         & $8,713$      \\
    total $\#$ reviews                 & $151,254$    \\
    avg. $\#$ reviews per item       & $17$         \\
    total $\#$ words                   & $16,666,135$  \\
    avg. number of words per review       & $110$         \\
    \hline
    \end{tabular}
    \caption{Statistics of the Amazon Grocery and Gourmet Food dataset.}
    \label{tab:statistics_amazon_food}
\end{table}

\begin{table*}[ht]
\centering
    \begin{tabular}{ccccccc}
    \hline
    \multirow{2}{*}{\thead{$n$\_factor}} & \multicolumn{6}{c}{\thead{ConvMF with different $\lambda$s}} \\
    \cline{2-7} 
     & \thead{$0.0$} & \thead{$0.4$}  & \thead{$0.8$} & \thead{$1.2$}  &  \thead{$1.6$}  &  \thead{$2.0$} \\
    \hline
    $6$ & $0.2850$  & $0.2932$ & $0.3508$ & $0.3863$ & $0.4043$ & $\bm{0.4161}$ \\
    $8$ & $0.2737$ & $0.2825$ & $0.3485$ & $0.3619$ & $0.3964$ & $\bm{0.4112}$ \\
    $10$ & $0.2634$ & $0.283$  & $0.3132$ & $0.3582$ & $0.3743$ & $\bm{0.3982}$ \\
    $12$ & $0.2424$ & $0.2804$ & $0.2937$ & $0.3599$ & $0.3722$ & $\bm{0.3808}$ \\
    \hline
    \end{tabular}
    \caption{Comparison of topic coherence (word embedding cosine similarity) for different entropy regularization coefficients ($\lambda$s), with $\lambda=0.0$ indicating the absence of entropy regularization terms. The highest-performing result is highlighted in bold.}
    \label{tab:word_similarity}
\end{table*}

\begin{table*}[ht]
\centering
    \begin{tabular}{ccccccccc}
    \hline
    \multirow{2}{*}{\thead{$n$\_factor}} & Offset & PMF & \multicolumn{6}{c}{\thead{ConvMF with different $\lambda$s}} \\
    \cline{2-9} 
     & / & / & \thead{$0.0$} & \thead{$0.4$}  & \thead{$0.8$} & \thead{$1.2$}  &  \thead{$1.6$}  &  \thead{$2.0$} \\
    \hline
    $6$ & $1.1722$ & $1.1467$ & $\bm{1.0632}$ & $1.0765$ & $1.0920$  & $1.0927$ & $1.0927$  & $1.0940$ \\
    $8$ & $1.1722$ & $1.1559$ & $\bm{1.0662}$ & $1.0797$ & $1.0915$ & $1.0902$ & $1.0982$  & $1.1047$ \\
    $10$ & $1.1722$ & $1.1607$ & $\bm{1.0735}$ & $1.0805$ & $1.0895$ & $1.0945$ & $1.0985$  & $1.1010$ \\
    $12$ & $1.1722$ & $1.1661$ & $\bm{1.0778}$ & $1.0852$ & $1.0902$ & $1.0985$ & $1.10325$ & $1.1117$ \\
    \hline
    \end{tabular}
    \caption{Comparison of rating prediction accuracy (RMSE) between base models and ConvMF under different entropy regularization coefficients ($\lambda$s), with $\lambda=0.0$ indicating the absence of entropy regularization terms. The highest-performing result is highlighted in bold.}
    \label{tab:rmse}
\end{table*}

\paragraph{Evaluation Metrics}
Our evaluation of the entropy regularization strategy on ConvMF focuses on two aspects.

The first aspect is the influence on the coherence of keywords within topics, which is quantified using cosine similarity on word embeddings. Cosine similarity between two embeddings is defined as in Equation \ref{cosine_sim}.
\begin{equation}\label{cosine_sim}
sim\left(\bm{w}_i, \bm{w}_j\right) = \dfrac{\bm{w}_i\cdot \bm{w}_j}{\|\bm{w}_i\|\|\bm{w}_j\|}
\end{equation}
\noindent 
For our experiments, we employ pre-trained GloVe embeddings \cite{pennington2014glove}. We compute all pairwise cosine similarities among the keywords within each topic and subsequently average these values to generate the final topic coherence metric.

The second aspect is the impact of this strategy on rating prediction accuracy, as measured by a classic metric: Root Mean Squared Error (RMSE).

\paragraph{Baseline Models}
The impact of entropy regularization on topic coherence is examined by comparing the cosine similarity metric of ConvMF with and without the entropy regularization term.

For rating prediction accuracy, we compare it not only with ConvMF without entropy regularization term but also introduce two baseline models for a more comprehensive analysis:
\begin{itemize}
\item Offset: This baseline model computes the average of all ratings in the training set and employs this mean value as its prediction for the ratings in the test set.
\item PMF: Probabilistic matrix factorization \cite{mnih2007probabilistic} is a collaborative filtering method that estimates user-item ratings by learning latent factor vectors in a probabilistic framework.
\end{itemize}

\paragraph{Implementation Details}
During the preprocessing phase, stop words and punctuation are eliminated since they can occasionally obstruct accurate topic extraction. We establish a uniform length of $64$ words for all textual reviews. To meet this criterion, we employ padding for reviews that do not reach this length. On the other hand, reviews that exceed this length are truncated to $64$ words. To achieve robust results, we assign values of $6$, $8$, $10$, and $12$ to the number of latent factors. The entropy regularization coefficient, $\lambda$, is assigned values of $0.4$, $0.8$, $1.2$, $1.6$, and $2.0$ to examine the influence of the coefficient's scale on the outcomes. The window size for the convolutional filters is fixed at $5$, meaning each activation value is derived from a $5$-gram within the textual reviews.

In our approach, we emphasize that only the textual reviews from the training set are harnessed for item embedding extraction throughout both the training and prediction phases, which means that the textual reviews within validation and test sets are strictly segregated and untouched. This practice mirrors real-world recommender systems where rating prediction occurs prior to user purchases, thus only historical reviews are available for use.

\paragraph{Results and Discussions}

Table \ref{tab:word_similarity} illustrates the impact of entropy regularization on topic coherence, as measured by Word2Vec cosine similarity. A noticeable trend in the results is the corresponding increase in word2vec cosine similarity as the value of $\lambda$ increases. This can be interpreted as the result of entropy regularization penalizing textual reviews with evenly distributed activation values, thereby forcing the model to concentrate more on a narrower range of reviews. When employing entropy regularization, the model pays greater attention to word embeddings that are effective in reducing information entropy, thereby leading the model to focus more on similar words that facilitate this goal. As the regularization coefficient, $\lambda$, increases, the effect of this strategy becomes more significant, leading to an increase in the Word2Vec cosine similarity. This implies that our entropy regularization strategy successfully intensifies the model's focus, thereby leading to enhanced coherence in the topics extracted.

\begin{figure*}[ht]
\begin{subfigure}{0.5\linewidth}
    \includegraphics[width=0.33\linewidth]{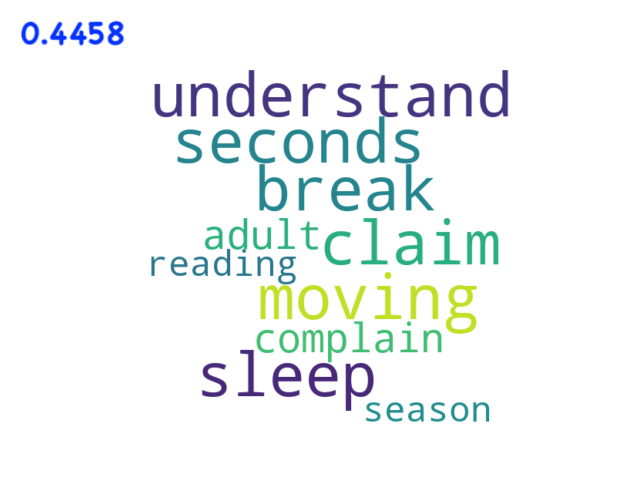}\hfill
    \includegraphics[width=0.33\linewidth]{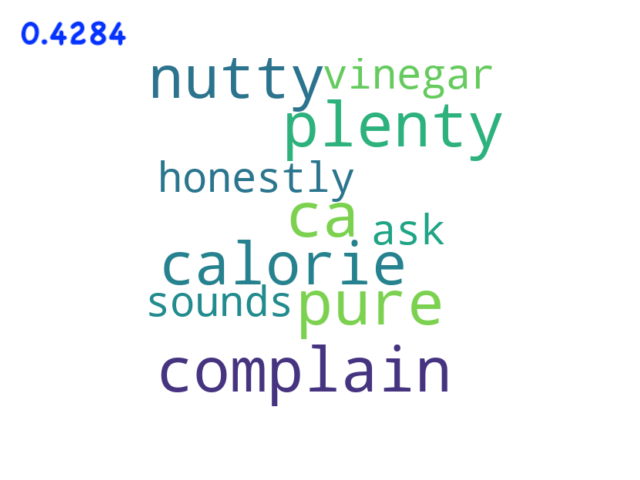}\hfill
    \includegraphics[width=0.33\linewidth]{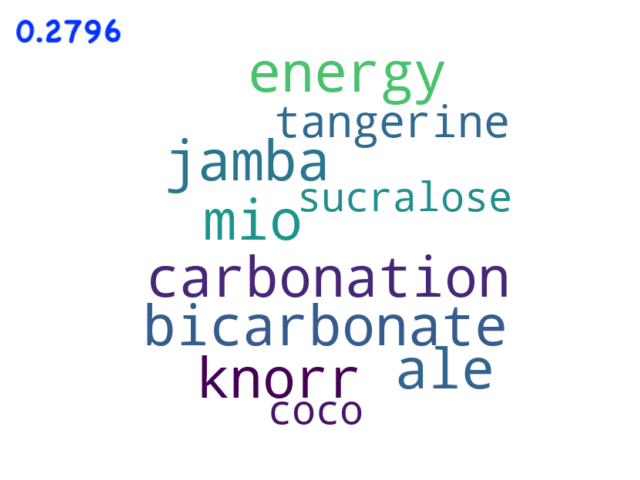}\hfill
    \includegraphics[width=0.33\linewidth]{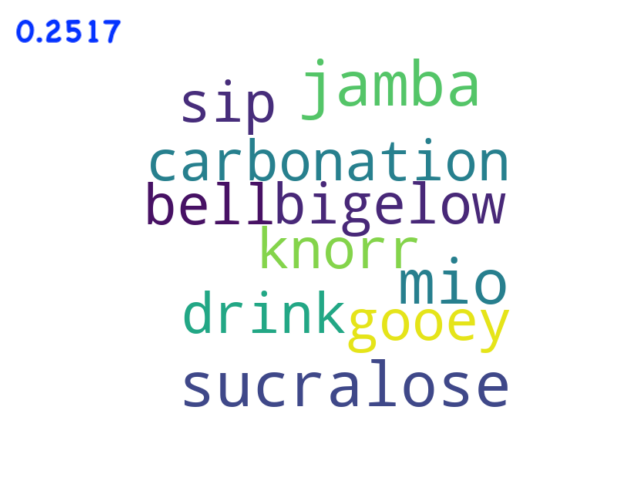}\hfill
    \includegraphics[width=0.33\linewidth]{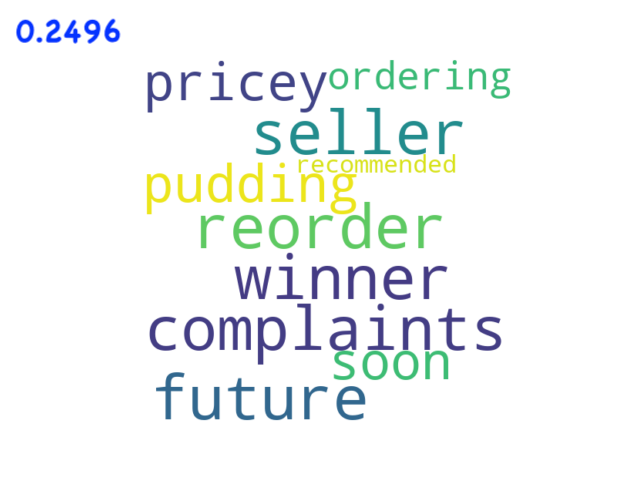}\hfill
    \includegraphics[width=0.33\linewidth]{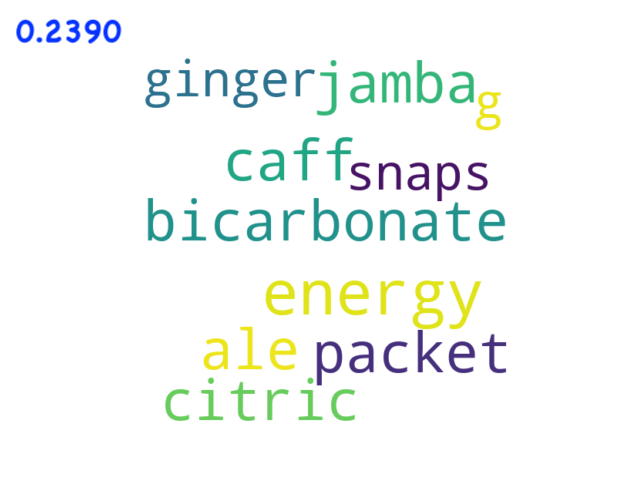}\hfill
    \includegraphics[width=0.33\linewidth]{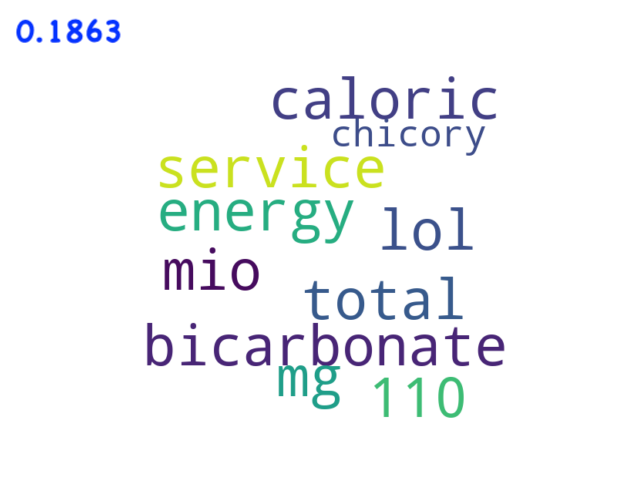}
    \includegraphics[width=0.33\linewidth]{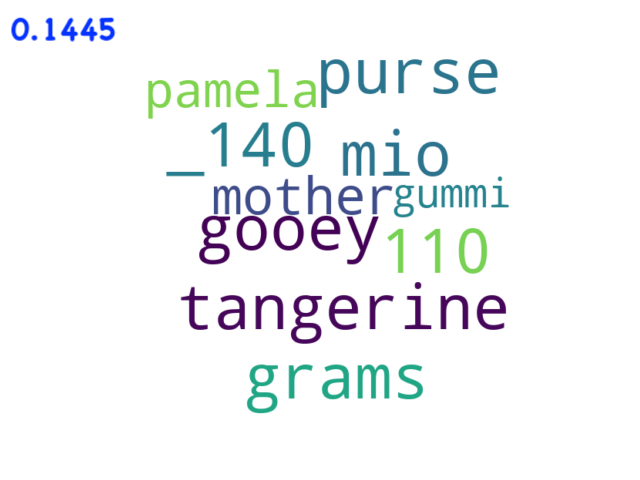}
    \subcaption{topic extraction without entropy regularization}
\end{subfigure}
\begin{subfigure}{0.5\linewidth}
    \includegraphics[width=0.33\linewidth]{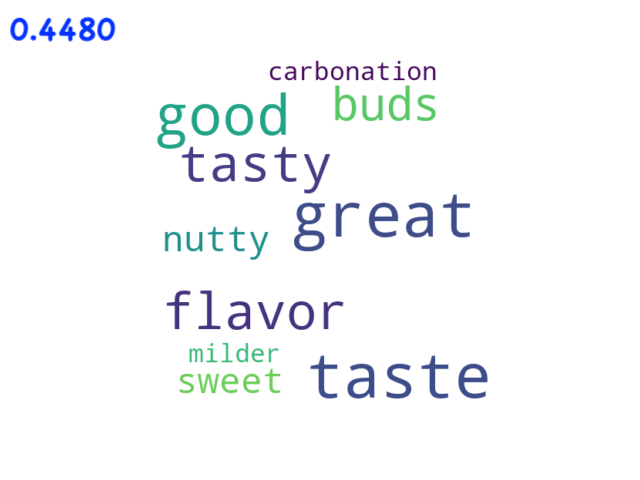}\hfill
    \includegraphics[width=0.33\linewidth]{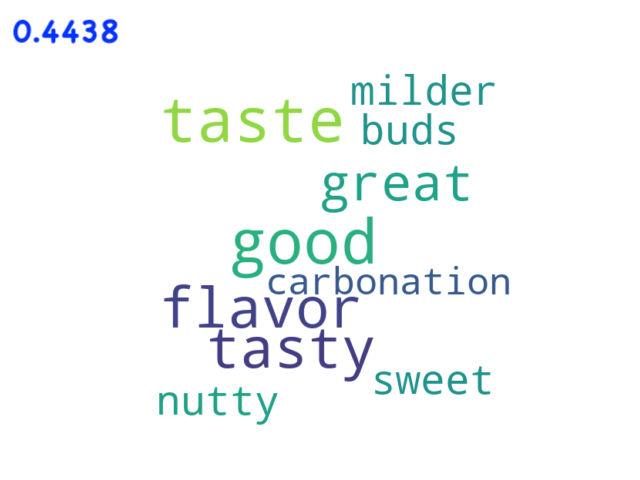}\hfill
    \includegraphics[width=0.33\linewidth]{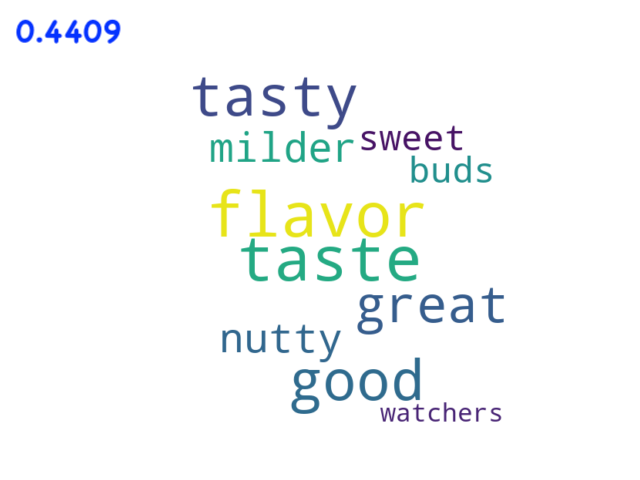}\hfill
    \includegraphics[width=0.33\linewidth]{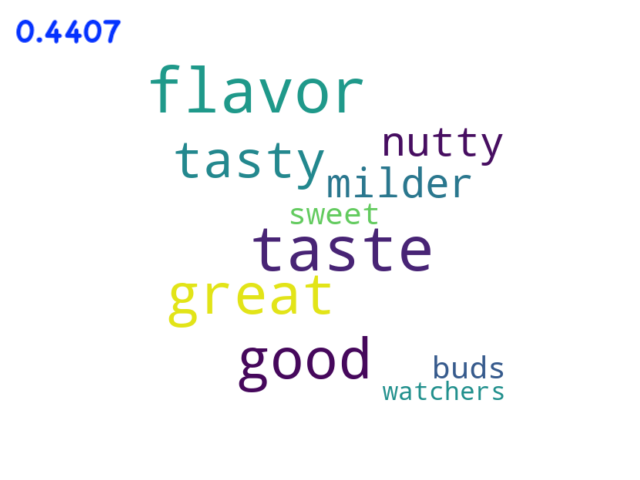}\hfill
    \includegraphics[width=0.33\linewidth]{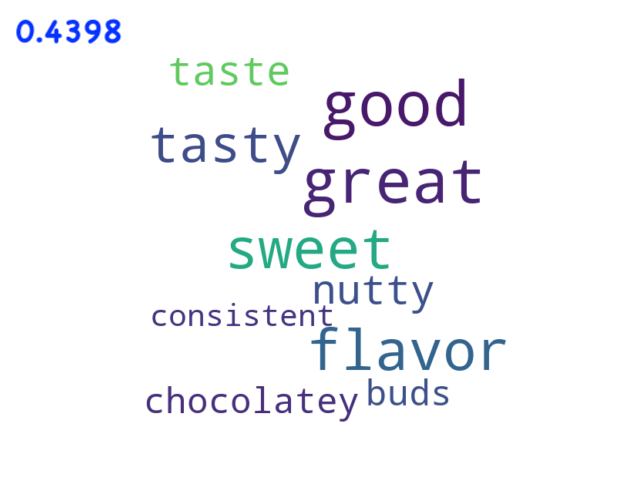}\hfill
    \includegraphics[width=0.33\linewidth]{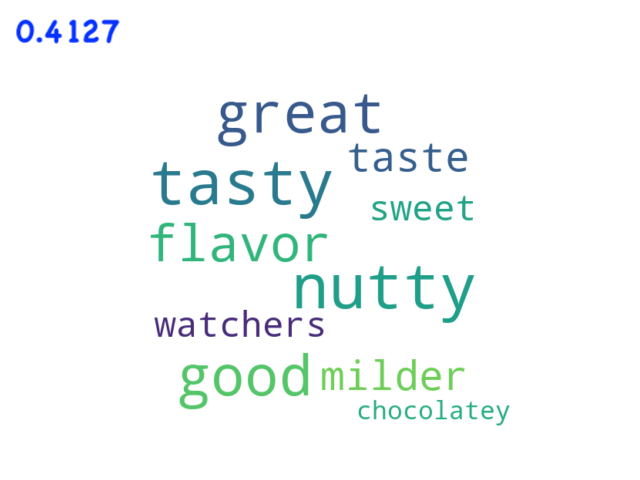}\hfill
    \includegraphics[width=0.33\linewidth]{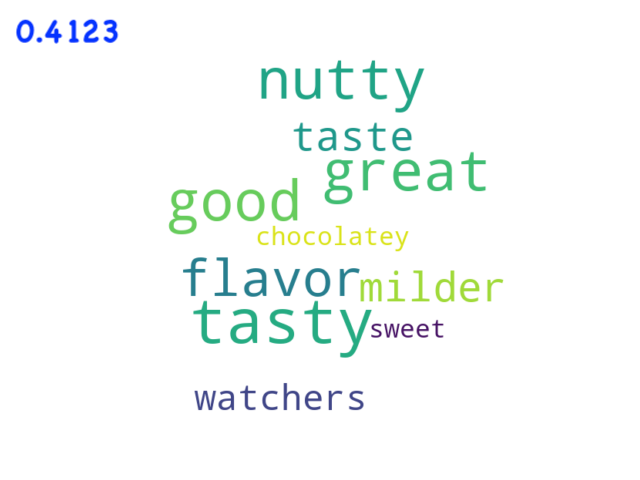}
    \includegraphics[width=0.33\linewidth]{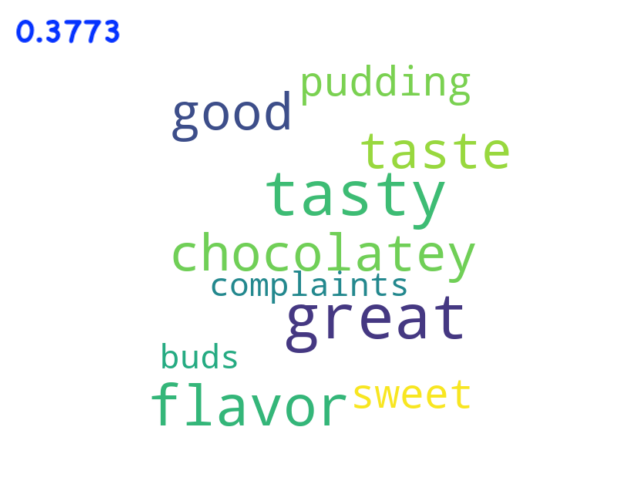}
    \subcaption{topic extraction with entropy regularization ($\lambda=2.0 $)}
\end{subfigure}\par\medskip
\caption{Illustration of extracted topic keywords. Here, the total number of latent factors is fixed at $8$. For each latent factor, we compute the word2vec cosine similarities, which are displayed at the top-left corner of each corresponding word cloud. The topics are sorted by this metric in descending order. Additionally, we calculate the average word2vec cosine similarities for all latent factors, yielding values of $0.2781$ and $0.4270$ respectively.}
\label{fig:topic_extraction}
\end{figure*}

To further elucidate the effectiveness of entropy regularization, we have visualized the extracted topics using word clouds, as depicted in Figure \ref{fig:topic_extraction}. For this visualization, we set the number of latent factors to $8$ and the coefficient of entropy regularization term to $2.0$. When using ConvMF without entropy regularization, it is evident that while a majority of the extracted topic keywords make sense, a subset appears less meaningful, such as "g", "ca", "110", "\_140", and "mg". However, upon the application of entropy regularization to the ConvMF framework, these less meaningful keywords significantly diminish. On the other hand, without entropy regularization, the coherence within a topic's keywords is compromised. For example, a topic may encompass keywords such as ['moving', 'understand', 'break', 'sleep', 'seconds', 'claim', 'complain', 'adult', 'reading', 'season'], which are not inherently related to one another. This lack of coherency is contrary to our expectations, as we aim for a concentrated semantic theme within each topic. Upon the implementation of entropy regularization, we observe a notable improvement in topic coherence. For example, a topic's keywords might include ['great', 'taste', 'good', 'flavor', 'tasty', 'buds', 'nutty', 'sweet', 'milder', 'carbonation'], most of which are positively connotative words that pertain to taste. In summary, the results clearly indicate that the topic keywords extracted by the model with entropy regularization are more meaningful and focused than those extracted without entropy regularization, contributing to a more interpretable model.

However, we have also observed a lack of topic diversity in ConvMF, and it appears that the entropy regularization strategy exacerbates this issue. For example, in the ConvMF without entropy regularization, the model tends to extract diverse keywords such as "mio", "bicarbonate", "jamba", and so on. However, when entropy regularization is employed, the model converges towards keywords like "taste", "great", "flavor" and "good". This decreased diversity in topic semantics indicates a homogenization of latent factors, which can negatively impact the overall performance of the recommender systems. 

Table \ref{tab:rmse} illustrates the effect of entropy regularization on rating prediction accuracy. The results suggest that entropy regularization slightly diminishes the accuracy of rating prediction. Our conjecture for this occurrence is that the entropy regularization, while improving topic coherence, may unintentionally compromise topic diversity. This reduction, in turn, slightly hampers the overall performance of the rating prediction task. However, it's important to note that this effect is rather subtle, and even with entropy regularization, the model continues to surpass both the Offset and PMF models in terms of predictive performance.

\section{Conclusion and Future Work}
This paper presents a novel approach to enhance topic extraction in recommender systems through the application of information entropy. By incorporating entropy regularization into ConvMF, we address the challenge of low coherence within topic keywords and achieve more accurate topic extraction. Our evaluation demonstrates that entropy regularization significantly improves topic coherence while maintaining competitive rating prediction accuracy. These findings highlight the potential of using information entropy in recommender systems and its broader applicability across various domains.

 However, while entropy regularization improves topic coherence, it does not necessarily improve the diversity of topics within the ConvMF model. As we observed, the model tends to concentrate on similar semantics, which could potentially limit the performance of the recommender systems. Therefore, future work could explore the following directions: 1) Investigate the employment of entropy regularization to other baseline models beyond ConvMF, aiming to see if they also witness an improvement in topic coherence without suffering a reduction in topic diversity. 2) Implement additional strategies within the ConvMF model to enhance topic diversity, possibly alongside entropy regularization, to maintain high topic coherence while promoting diverse semantic themes.

\section*{Acknowledgments}
This research was supported by Science Foundation Ireland (SFI) under Grant Number SFI/12/RC/2289.

\bibliographystyle{abbrv}
\setcitestyle{citesep={,}}
\bibliography{main}

\end{document}